\documentclass{ieeeaccess}
\usepackage{cite}
\usepackage{amsmath,amssymb,amsfonts}
\usepackage{algorithmic}
\usepackage{graphicx}
\usepackage{textcomp}
\usepackage{mathtools}
\usepackage{amsmath}
\usepackage{amssymb}
\DeclarePairedDelimiter\floor{\lfloor}{\rfloor}

\usepackage{multirow}
\usepackage{booktabs}
\usepackage{algorithm}
\usepackage[algo2e]{algorithm2e}

\newcommand{\etal}{\textit{et al}.}
\newcommand{\ie}{\textit{i}.\textit{e}.}
\usepackage{pifont}
\newcommand{\cmark}{\ding{51}}%

\def\BibTeX{{\rm B\kern-.05em{\sc i\kern-.025em b}\kern-.08em
    T\kern-.1667em\lower.7ex\hbox{E}\kern-.125emX}}
\begin{document}
\history{Date of publication August 19, 2020, date of current version August 31, 2020.}
\doi{10.1109/ACCESS.2020.3017881}

\title{TiVGAN: Text to Image to Video Generation with Step-by-Step Evolutionary Generator}
\author{\uppercase{DOYEON KIM}\authorrefmark{*} \IEEEmembership{Member, IEEE},
\uppercase{DONGGYU JOO\authorrefmark{*}, and Junmo kim}
\IEEEmembership{Member, IEEE}}

\address{School of Electrical Engineering, Korea Advanced Institute of Science and Technology, Daejeon 34141, South Korea}

\tfootnote{*Doyeon Kim and Donggyu Joo contributed equally to this work.}

\markboth
{Kim \headeretal: TiVGAN: Text to Image to Video Generation with Step-by-Step Evolutionary Generator}
{Kim \headeretal: TiVGAN: Text to Image to Video Generation with Step-by-Step Evolutionary Generator}

\corresp{Contact: doyeon\_kim@kaist.ac.kr, jdg105@kaist.ac.kr \\
Corresponding author: Junmo Kim (e-mail: junmo.kim@kaist.ac.kr).}

\begin{abstract}
  Advances in technology have led to the development of methods that can create desired visual multimedia. 
  In particular, image generation using deep learning has been extensively studied across diverse fields. 
  In comparison, video generation, especially on conditional inputs, remains a challenging and less explored area. 
  To narrow this gap, we aim to train our model to produce a video corresponding to a given text description. 
  We propose a novel training framework, Text-to-Image-to-Video Generative Adversarial Network (TiVGAN), which evolves frame-by-frame and finally produces a full-length video. 
  In the first phase, we focus on creating a high-quality single video frame while learning the relationship between the text and an image. 
  As the steps proceed, our model is trained gradually on more number of consecutive frames.
  This step-by-step learning process helps stabilize the training and enables the creation of high-resolution video based on conditional text descriptions. 
  Qualitative and quantitative experimental results on various datasets demonstrate the effectiveness of the proposed method.
\end{abstract}


\titlepgskip=-15pt

\maketitle

\section{Introduction}
\label{sec:introduction}
\PARstart{I}{n} the last few years, there has been intensive research on generative models. In particular, the recent developments of variational auto encoders(VAEs)~\cite{kingma2013auto} and generative adversarial networks(GANs)~\cite{goodfellow2014generative} represent the forefront of rapid, abundant, and high-quality progress. Further, since the deep convolutional GAN~\cite{radford2015unsupervised}, which employs a convolutional neural network(CNN), succeeded in generating a realistic output using a GAN framework, many studies have reported impressive results using deep networks. It is now possible to produce photo-realistic images that are difficult to discriminate from real images, even for humans~\cite{karras2017progressive}.

\Figure[!t](topskip=0pt, botskip=0pt, midskip=0pt)[width=1\linewidth]{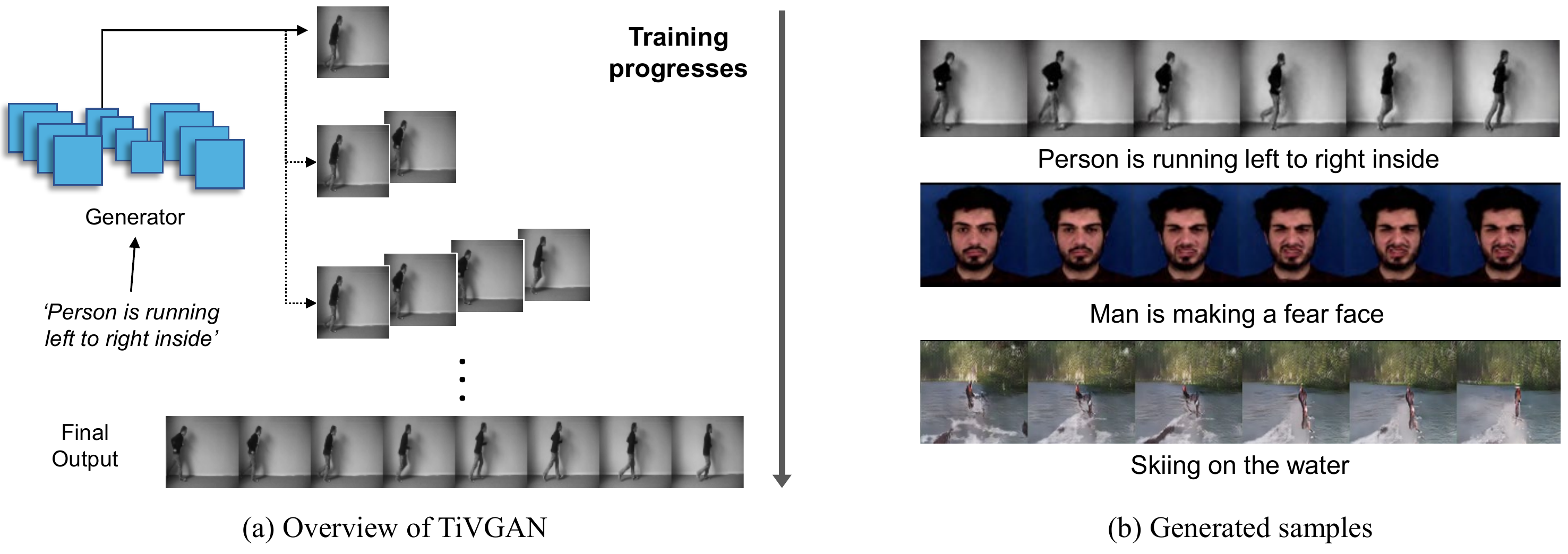}
{\textbf{Simple overview of our TiVGAN framework and generated videos.} (a) The generator starts with producing a single frame and gradually evolves to create longer frames with the given text. (b) Generated sample videos using our framework TiVGAN.\label{fig:overview}}

However, the number of studies on video generation is significantly lower than those on image generation, because the video generation is a considerably more challenging task than image generation. Image creation is only concerned with the completeness of a single frame, whereas videos also need to consider the connectivity between frames. Even if each image has good quality, well-crafted videos cannot be generated if the continuity between adjacent frames is not guaranteed. In addition, nearly all public video datasets are extremely diverse and unaligned, thereby further complicating the video generation process.

Conditional video generation has received little research attention, whereas the generation of conditional output for a wide variety of inputs has been widely studied in the image generation field~\cite{chen2016infogan, mirza2014conditional, odena2017conditional}. For example, a simple one-hot vector can be used as a control code to manipulate the attributes of a resulting image~\cite{chen2016infogan}, and there is also a network that creates photo-realistic images corresponding to a given text~\cite{zhang2017stackgan}. However, studies regarding text-to-video generation are lacking and are generally performed on a low resolution compared to text-to-image generation. Therefore, to broaden the field of video generation, we focused on generating a conditional video that has not yet been investigated thoroughly in this domain. This study introduces a new scheme for text-to-video generation tasks with GANs. 

We propose a novel network that generates a video corresponding to a given description. The learning framework of our network is established on the basic concept that connected frames of a video have substantial continuity. If we can create one high-quality video frame, it will be easier to create a linked frame because they are related. Rather than first finding a mapping function between the text and all video frames, we train our network with respect to one image and gradually extend it to longer frames (Figure~\ref{fig:overview}). Our model call this scheme as TiVGAN, which stands for Text-to-Image-to-Video Generative Adversarial Network framework. In the process of progressively learning to generate a large number of adjacent frames, TiVGAN can learn to create long consecutive scenes. Our extensive experimental results show that our model not only produces an accurate video for a given text but also produces qualitatively and quantitatively sharper and better results than those presented in other comparable works.

\section{Related Works}

Generative image models have been studied actively in the past few years. Kingma~\etal~\cite{kingma2013auto}  suggested a re-parametrization trick to derive a variational lower limit of data likelihood. GAN shows promising results with the use of adversarial training between discriminators and generators. The discriminator is trained to distinguish between fake and real distribution, and the generator attempts to create realistic data to deceive the discriminator. Since then, there have been creation tasks for various datasets, such as human faces, furniture, animals, and others~\cite{berthelot2017began,  miyato2018spectral, radford2015unsupervised}.

Following the research on the generation of images based on GANs, studies regarding conditional GANs have also been published with various kinds of conditional inputs. InfoGAN~\cite{chen2016infogan} uses a one-dimensional vector as a condition to control output image by concatenating the code into input noise. Maximizing the mutual information between the code and the generated image enables the network to learn interpretable representations. Furthermore, Reed~\etal~\cite{reed2016generative} demonstrated networks that can create text-based images by learning text feature representations and using them to synthesize images. StackGAN~\cite{zhang2017stackgan} extends this structure to stage-$1$ and stage-$2$, which enables the generation of $256\times256$ photo-realistic images after the generation of low-resolution images. Moreover, there have been many studies on the whole-image translations, such as image domain transfer~\cite{isola2017image, yi2017dualgan, zhu2017unpaired} and image manipulation~\cite{joo2018generating, shen2017learning, tran2017disentangled}.

In contrast, very few experiments have been conducted on video synthesis. Vondrick~\etal~\cite{vondrick2016generating} untangled the background and foreground of the scenes with two streams using 2D spatial convolution and 3D spatio-temporal convolutions for each scene. TGAN~\cite{saito2017temporal} exploits two different generators for temporal vector sampling and multiple frame creation based on the acquired vectors. The developers of MoCoGAN~\cite{tulyakov2018mocogan} suggested the decomposition of motion and content space for effective video generation. They used a recurrent neural network for sampling from a motion subspace and concatenated the sampled features with the content vector to generate continuous frames. Our goal of text-to-video generation, on the other hand, has rarely been attempted. Li~\etal~\cite{li2018video} used a conditional VAE to generate a `gist', which refers to the video background color and object layer; the video content and motion are created based on the gist and text. Pan~\etal~\cite{pan2017create} proposed a new architecture for the text-to-video task by using 3D convolutions on their network and different types of losses. Balaji~\etal~\cite{balaji2019conditional} suggested a multi-scale text conditioning scheme with GAN to generate desired video frames with the given text. However, despite these examples, the number of studies on text-to-video remains small. Therefore, we present a new architecture suitable for video generation conditioned on the text description.

\Figure[!t](topskip=0pt, botskip=0pt, midskip=0pt)[width=1\linewidth]{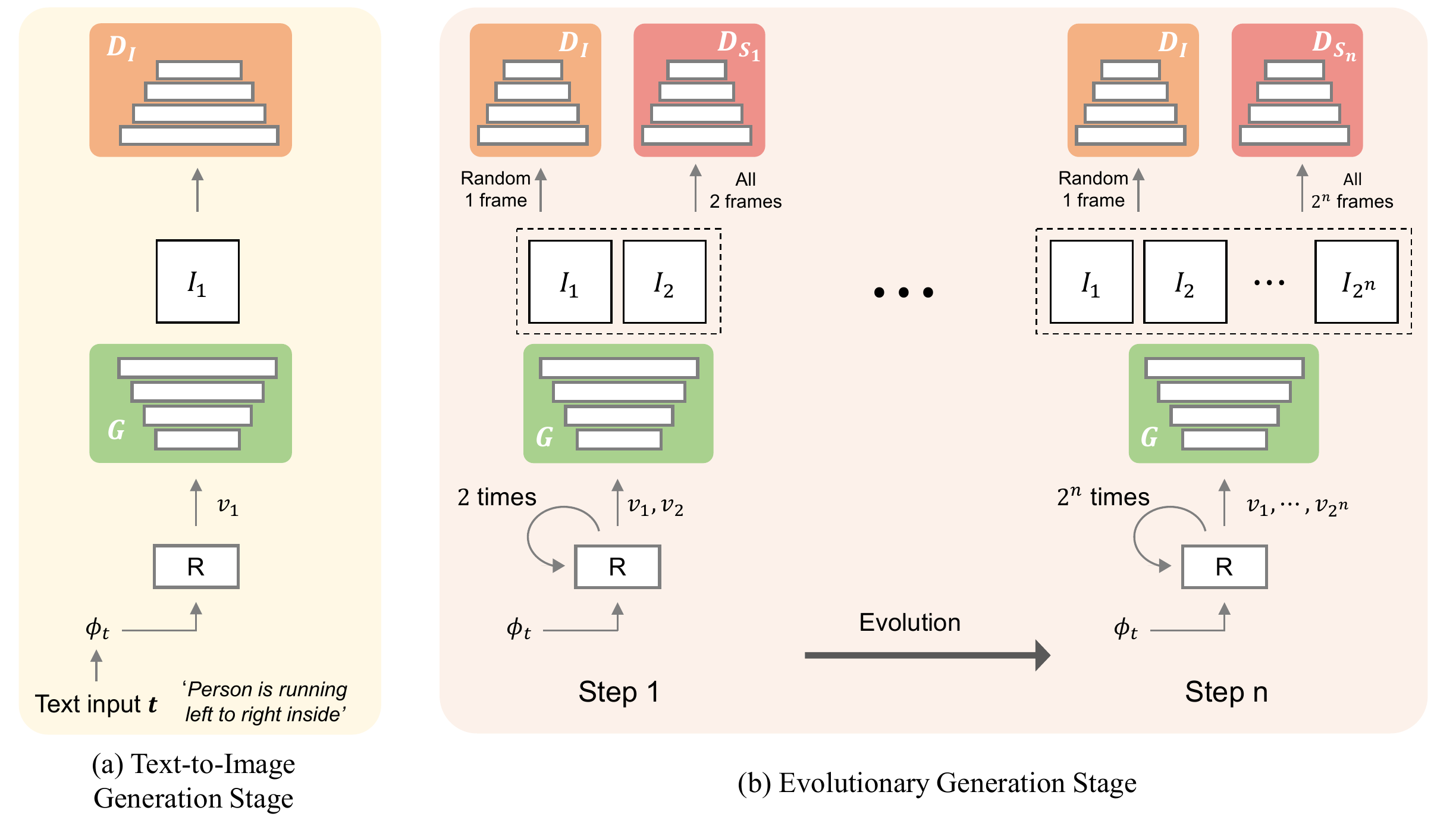}
{\textbf{Full architecture of our proposed network, TiVGAN, and the training stages.} We first start with training for generating a single image at the text-to-image generation stage, and we make consecutive frames in an evolutionary way through further stages. Although the text-to-image generation stage only uses an image discriminator $D_I$, the evolutionary generation stage uses both an image discriminator $D_I$ and a step-discriminator $D_{S}$.\label{fig:main_figure}}

\section{Method}
\label{method}

As we have emphasized in the above sections, GANs have been proven for their ability to create sharp images. From this point of view, starting our text-to-video network training with a text-to-image stage may result in more effective video generation. Based on these ideas, we decomposed the training process into two stages: \textit{Text-to-Image Generation} and \textit{Evolutionary Generation}. The overall flow of our proposed architecture is described in Figure~\ref{fig:main_figure}. We begin with the learning of text-to-single image generation, gradually increase the number of produced images, and repeat this training process until the desired video length is achieved. This is our key paradigm. These two stages will be described in detail in subsections~\ref{sec:texttoimage} and~\ref{sec:evolutionary_generator}, followed by explanations of the techniques we used to stabilize the learning.

\subsection{Text-to-Image Generation}
\label{sec:texttoimage}
We aim to generate a realistic fake video $V_{F}=(I_1, I_2, \cdots, I_{2^n})$ that matches with the given text description $t$ using a recurrent unit $R$ and a generator $G$ where $n \in \mathbb{N}$ and $I_i$ is each frame of video $V_{F}$. At this stage, we only focus on the \textit{text-to-image generation} task without considering the image sequence. Then, our goal is simplified to the generation of the single realistic image $I_1$ from $t$. 

To train the model with text, we must first transform the text string into an encoded feature vector. We adopted the pre-trained skip-thoughts vector network~\cite{kiros2015skip} to encode text $t$ into a $4,800$-dimensional vector. Since the encoded vector is high-dimensional, we used principal component analysis (PCA) to derive meaningful features and reduce its dimensionality. We defined this embedded vector as $\phi_t$.

We start with a single GRU cell $R$ and a generator $G$ to create one frame. Given the embedded text vector $\phi_t$, the recurrent unit $R$ outputs the vector $v_1=R(z_0, (\phi_t, z_1))$, where $z_0$ and $z_1$ are random noises from $\mathcal{N}(0,1)$. The noise $z_0$ is a initial hidden state, and $(\phi_t, z_1)$ is an input of $R$.
This $v_1$ from $R$ is the source input vector for $G$, and it creates the resulting image $I_1 = G(v_1)$ with the same size as the real frame. To ensure that $I_1$ matches with the provided text description $t$ and follows the distribution of real data, we trained the generator $G$ and the image discriminator $D_I$ adversarially using a GAN framework with slightly modified losses consisting of real, wrong, and fake pair as similar to those used by Reed \etal~\cite{reed2016generative}. Overall losses will be explained in Section~\ref{sec:training_procedure}.

When training the image discriminator $D_I$, the real image $X_i$ is randomly selected from $2^n$ frames of the real video $V_R$. Therefore, $G$ and $R$ aim not only to create the corresponding image for the given text but also to model the image distribution of the frames in the given video dataset. After the training of the text-to-image generation, $G$ can generate various images if some appropriate input vector is received. Therefore, if $R$ provides a meaningful sequence of vectors, $G$ can easily generate consecutive frames, which leads to realistic video generation. $R$ could act as an instructor that teaches $G$ to synthesize consecutive frames. This initial stage is the basis of the whole training process and plays a significant role in generating a series of frames. 

\subsection{Evolutionary Generation}

\label{sec:evolutionary_generator}
The \textit{evolutionary generation} stage begins after the completion of the initial stage training described in the above section. We now create a series of consecutive frames based on the model trained in the previous stage. Since $G$ has the ability to generate various frames and $R$ is a recurrent unit that can output a series of meaningful vectors, the extension of the text-to-image generation can lead to successful consecutive frames generation. 

This stage consists of the process from step $1$ to $n$, and the goal of each step $m \in \{1,2,\cdots n\}$ is to generate $2^m$ consecutive frames stably. 
As the step proceeds, the number of created frames increases, and we can finally reach the video-level generation we desired. 
In each step $m$, $2^m$ images can be obtained by iterative operations of $R$ and $G$ learned in the previous stage. 
Let's look at an example of the generation in step $1$. 
After the text-to-image generation, we forward $R$ once more with $v_1$ as a hidden state and the ($\phi_t$, $z_2$) as an input where $z_2$ is randomly sampled noise from $\mathcal{N}(0,1)$. 
Then the next latent vector $v_2=R(v_1, (\phi_t, z_2))$ is obtained from $R$. 
This vector $v_2$ is again delivered through the same $G$ to create another image, $I_2$. 

Unlike the text-to-image generation stage, the temporal consistency of the generated frames should be managed in this stage. At each step $m$, $m^{th}$ \textit{step-discriminator} $D_{S_m}$ is newly added to discriminate the sequence of $2^m$ frames. $D_{S_m}$ receives the fake input $(I_1, I_2, \cdots, I_{2^m}, \phi_t)$ and the real input $(X_1$ $, X_2, \cdots, X_{2^m}, \phi_t)$ where $(X_1, X_2, \cdots, X_{2^m})$ are $2^m$ randomly selected connected frames from the real video $V_R$.  
Since the real input has temporal information (images are concatenated with original order), the fake input should be generated to have temporal information correctly.
After this training step converges, we move on to the further step $m+1$. Then, $D_{S_m}$ is removed, and training proceeds with a new step-discriminator $D_{S_{m+1}}$.

To summarize, we use different step-discriminator $D_{S_1}, \cdots, D_{S_{n}}$ on individual $1, 2, \cdots, n$ steps. For $m \in \{1,2, \cdots, n\}$, $R$ repeats $2^{m}$ times, and $G$ generates a sequence of $2^m$ images from input vector $v_1, v_2, \cdots, v_{2^{m}}$. $D_{S_{m}}$ is initialized at the beginning of step $m$ to discriminate the $2^m$ images, and it is only used for the step $m$. 
Meanwhile, $D_I$ is used through all steps $1,2, \cdots, n$ to maintain high-quality image output. This step-by-step training converges quickly and stably. 
When all the training finishes, we can easily generate the images by forwarding single $R$ and $G$ for $2^n$ times, and it forms the video with $2^n$ lengths. The detailed training procedure and algorithm is described in Section~\ref{sec:training_procedure} and Algorithm~\ref{sec:al1}.\\

\Figure[!t](topskip=0pt, botskip=0pt, midskip=0pt)[width=0.99\linewidth]{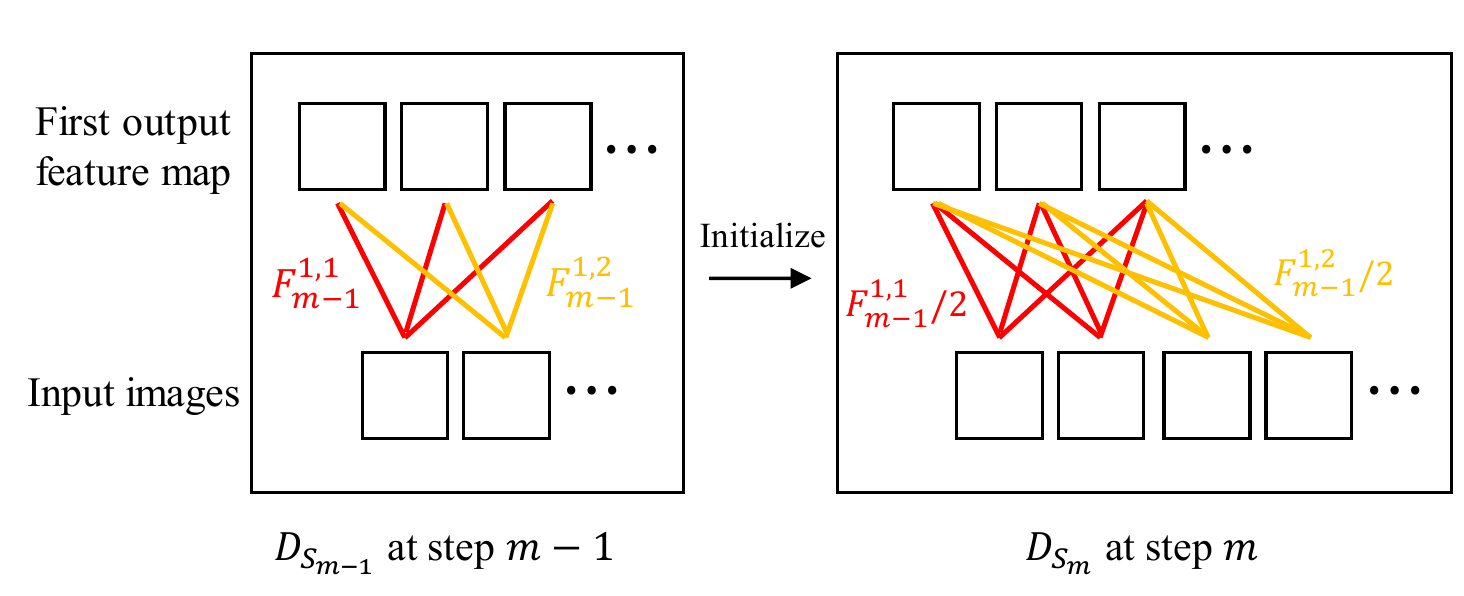}
{\textbf{Illustration of the first convolution layer of step-discriminators when step changes.} At the beginning of step $m$ of evolutionary generation stage, step-discriminator $D_{S_{m}}$ is initialized using $D_{S_{{m-1}}}$.\label{fig:init}}

\subsection {Step-discriminator Initialization}
\label{sec:step_discriminator}

Each step-discriminator $D_{S}\in\{D_{S_1}, \cdots, D_{S_{n}}\}$ is newly added at the beginning of every step of the evolutionary generation stage. Since $D_I$ is already trained through the steps, there could be an imbalance in the training progress of discriminators $D_I$ and $D_S$.
Therefore, simply applying the adversarial learning of $D_I$ and $D_S$ can even disrupt the learning of $G$. 
To prevent this, we present an advanced scheme which initializes the $D_S$ to a better state rather than noise.

For all $m \in \{1,2, \cdots, n\}$, the weight of $D_{S_{m}}$ is initialized with the previous step-discriminator $D_{S_{m-1}}$. Image discriminator $D_I$ is used for initialization of the first step-discriminator $D_{S_1}$.
We designed all the step-discriminators to have the same architecture except for the number of input channels in the first layer. 
The only difference is that $D_{S_{m}}$ receives $2 ^ m$ images as input, while $D_{S_{{m-1}}}$ receives $2 ^ {m-1}$ images. 
Let $F_{m}^{l, k}$ be the weights connected to the $k^{th}$ input channel of the $l^{th}$ layer of step-discriminator $D_{S_{m}}$. Then, our step-discriminator initialization can be defined as:

\begin{equation}
  F_{m}^{l, k}=\begin{cases}
    F_{m-1}^{l, \floor{\frac{k+1}{2}}}/2 & \text{ if \textit{l} = 1}.\\
    F_{m-1}^{l,k} & \text{ if \textit{l}  $\neq$ 1}.
  \end{cases}
\end{equation}

All layers except the first layer are initialized with same weight of the step-discriminator $D_{S_{m-1}}$. However, there is a slight variation only in the first layer, which is that $F_{m}^{1, 2i-1}$ and $F_{m}^{1, 2i}$ are initialized to $F_{m-1}^{1, i}/2$ for all $i=1,\cdots, 2^{m-1}$. This is illustrated in Figure~\ref{fig:init}.
After the evolution in step $m-1$ to step $m$, the appearance of the generated images within each step should be similar, but the number of the resulting images are twice as long. To retain the output of the discriminator when the step changes, we divide the weight value by $2$. 
We believe that this initialization technique can assist the training of our framework to maintain stability even with sudden step changes. The effect of this method will be discussed further in the results section.

\subsection{Independent Samples Pairing}

One of the main failures in training GAN is mode collapse. 
To mitigate this phenomenon, \textit{independent samples pairing} (ISP) is applied in our training process. Our model produces two output images ${I_k}^a$ and ${I_k}^b$ with one fixed text description $t$ and two different randomly sampled input noises $z_k^a$ and $z_k^b \in \mathcal{N}(0,1)$ when generating the $k^{th}$ frame ($k=1,2, \cdots,2^n$). These two independently generated images are paired by concatenation in the channel dimension and form the fake pair. To make the generator create various examples corresponding to the same text $t$, we train the discriminator to distinguish between $({I_k}^a, {I_k}^b, \phi_t)$ and $(X^a, X^b, \phi_t)$, where $X^a$ and $X^b$ are two real dissimilar images associated with the same text description. 
Since $X^a$ and $X^b$ are dissimilar, if a mode collapsed generator generates very similar $I_k^a$ and $I_k^b$, it will be easily detected as fake by the discriminator. Thus, the generator attempts to create different images with the same $t$ to deceive the discriminator. This independent samples pairing technique is only used for image discriminator $D_I$. 
\label{sec:isp}

\subsection{Training Procedures}
\label{sec:training_procedure}
We retain the adversarial training framework of the generator and discriminator. Unlike in a conventional GAN, however, we include a slight perturbation by having two branches on the discriminator similar to ~\cite{zhang2017stackgan}. At first, the discriminator passes several convolution layers to acquire a high-level feature map. Then, one branch calculates the text-image match loss by concatenation with $\phi_t$, and the other branch performs patch discrimination without text concatenation. The operations of these two independent branches ensure that the image matches well with the text while improving the image quality. We use three kinds of text-image losses to train the model for text matching, the same as those used by Reed \etal~\cite{reed2016generative}. The losses are obtained by a real pair ($X$, $\phi_t$), fake pair ($I$, $\phi_t$), and wrong pair ($X$, $\hat{\phi_t}$), where $\hat{\phi_t}$ is one of the embedded text vector that is not identical to $\phi_t$.

The overall loss for $G$, $R$, $D_I$, and $D_S$ are as follows:
\begin{equation}
\label{eq:total_loss}
\small
\begin{split}
L_{obj}(G,R, D_I, D_S) &= L_I(G, R, D_I) + L_S(G, R, D_S),
\end{split}
\end{equation}
where
\begin{footnotesize}
\begin{multline}
\label{eq:loss_I}
L_I(G, R, D_I)  = \sum_{i}^{}(\log(D_{I}(X^{i})) + \log(D_{I}(X^{i}, \phi_t)) +\\ 
  \log(1-D_{I}(I^{i})) + \log(1-D_{I}(I^{i}, \phi_t)) + \log(1-D_{I}(X^{i}, \hat{\phi_t})))
\end{multline}
\end{footnotesize}
and
\begin{footnotesize}
\begin{multline}
\label{eq:loss_S}
L_S(G, R, D_S)  = \sum_{i}^{}(\log(D_{S}(X^{i, S})) + \log(D_{S}(X^{i,S}, \phi_t)) +\\ 
  \log(1-D_{S}(I^{i,S})) + \log(1-D_{S}(I^{i,S}, \phi_t)) + \log(1-D_{S}(X^{i,S}, \hat{\phi_t}))).
\end{multline}
\end{footnotesize}

\begin{algorithm}[H]
\SetAlgoLined
 \KwData{Text \textit{t}, corresponding real video $V_R=(X_1, X_2, \cdots, X_{2^n})$}
 {\textbf{Network: }Generator $G$, GRU $R$, Discriminators $D_I$, $D_{S_1}$, $\cdots$, $D_{S_{n}}$}
 \\
 \textbf{(I) Text-to-Image generation stage:} Generate a single image and train $G, R, D_I$ \\
 \While{not converged}{
 \vspace*{0.1cm}{Get $z_0, z_1 \in \mathcal{N}(0,1)$, generate ${I_1}=G(R(z_0, (\phi_t, z_1))$}\\
 \vspace*{0.1cm}{Randomly choose one real image $X$ from $V_R$} \\
 \vspace*{0.1cm}Update the $G, R$ $\leftarrow$ minimizes loss via Eq.~\ref{eq:loss_I} \\
 \vspace*{0.1cm}Update the $D_I$ $\leftarrow$ maximizes loss via Eq.~\ref{eq:loss_I} 
  with \textit{independent samples pairing}} 
 \textbf{(II) Evolutionary Generation stage.} 
 \\
 At each step-$m$, generate $2^{m}$ images and train $G, R, D_I, D_{S_{m}}$\\
 \For{$m\rightarrow 1$ \KwTo $n$}{
 \While{not converged}{
 \vspace*{0.1cm}{Get $z_1$,  $\cdots$, $z_{2^m} \in \mathcal{N}(0,1)$, generate $I_1$,  $\cdots$, $I_{2^m}$} by repeating $R$ and $G$ \\
 \vspace*{0.1cm}{Randomly choose $2^m$ consecutive images $X_1$, $\cdots$, $X_{2^m}$ from $V_R$} \\
\vspace*{0.1cm}{Update the $G, R$ $\leftarrow$ minimizes loss via Eq.~\ref{eq:loss_I},~\ref{eq:loss_S}} \\
 \vspace*{0.1cm}{Update the $D_I$ $\leftarrow$ maximizes loss via Eq.~\ref{eq:loss_I} with \textit{independent samples pairing}} \\
 \vspace*{0.1cm}{Update the $D_{S_{m}}$ $\leftarrow$ maximizes loss via Eq.~\ref{eq:loss_S}} 
  }
  }
 \caption{Training procedures}
 \label{sec:al1}
\end{algorithm}

$X^{i}$ and $I^i$ are randomly selected frames from real and fake videos, respectively. $D_I(I^{i})$ is an image loss without text conditioning, and $D_I(I^{i}, \phi_t)$ is the pair loss with text conditioning. At evolutionary step $m$, $D_S$ represents $D_{S_{m}}$. Here, $X^{i,S}$ is a $2^m$ consecutive images set randomly selected from the real video, and $I^{i,S}$ is generated fake images set with $2^m$ images. They are concatenated in channel dimensions. $L_S$ is only used at the evolutionary generation stage, and $L_I$ is used through all processes. Our overall training algorithm is described in Algorithm~\ref{sec:al1}.

\Figure[!t](topskip=0pt, botskip=0pt, midskip=0pt)[width=1\linewidth]{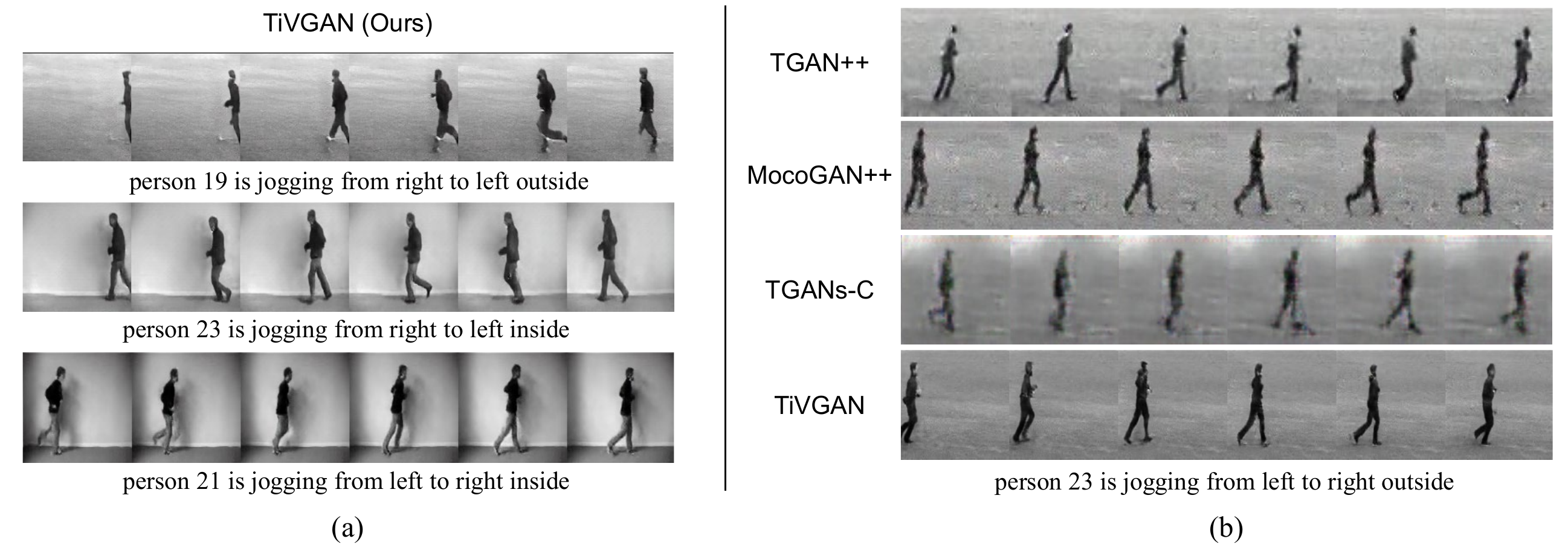}
{\textbf{Qualitative results of the models trained on KTH dataset. } (a) Our generation results. (b) Comparative results with previous works.\label{fig:KTH}}

\Figure[!t](topskip=0pt, botskip=0pt, midskip=0pt)[width=1\linewidth]{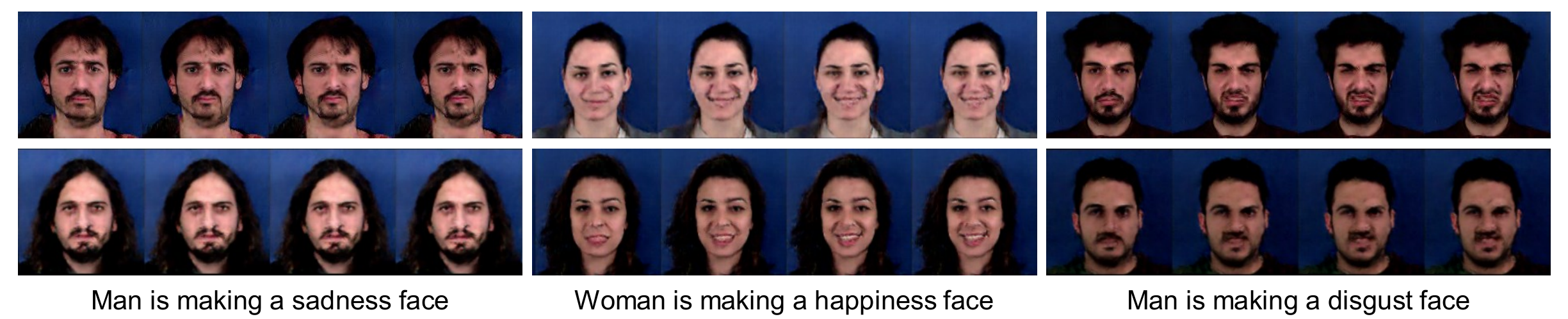}
{\textbf{Qualitative results for MUG dataset. } The images above and below are parts of generated video frames created from the same text description. Generated video frames are well matched with given text description.\label{fig:mg_qual}}

\section{Experiments}
\label{sec:experiments}
To prove the effectiveness of our proposed method, we conduct several experiments on three diverse datasets: KTH Action, MUG, and Kinetics.  To compare with the previous text-to-video method, we reproduced~\cite{pan2017create} and make experiments on three datasets. Since other methods ~\cite{li2018video, balaji2019conditional} have experimented on the Kinetics dataset, we directly compare results for the Kinetics. Due to the lack of published papers in the text-to-video area, we also employ several other existing video generation methods (TGAN~\cite{saito2017temporal}, MoCoGAN~\cite{tulyakov2018mocogan}), and trained on each dataset using the same settings. For a fair comparison, we attempted to balance our model and video generation model by adding text conditioning to each method to yield TGAN++ and MoCoGAN++. 

For all experiments, we used $n=4$ steps, which implies that we generated a 16-frame video. The detailed structure of the network is given in the supplementary material. In the training of the text-to-image generation stage, 30k iterations are performed, and in the evolutionary generation stage, we perform 15k iterations for each step. For PCA, we reduce the dimension of the vector to $60$, \ie, $\phi_t\in \mathbb{R}^{60}$.

\subsection {KTH Action}

The KTH Action dataset~\cite{laptev2004recognizing} contains six types of human actions: walking, jogging, running, boxing, hand waving, and hand clapping. Of these, we use the jogging class. In each clip, a man is jogging from left to right or right to left on two backgrounds. We extracted 16 frames from each video sequence and reshaped it into $128\times128$. For training, a total of 200 videos (3,200 frames) are used. 
Our qualitative results are shown in Figure~\ref{fig:KTH}. We can see that the person in the generated video is moving exactly as described in the text description while maintaining the image quality with high resolution.

\Figure[!t](topskip=0pt, botskip=0pt, midskip=0pt)[width=1\linewidth]{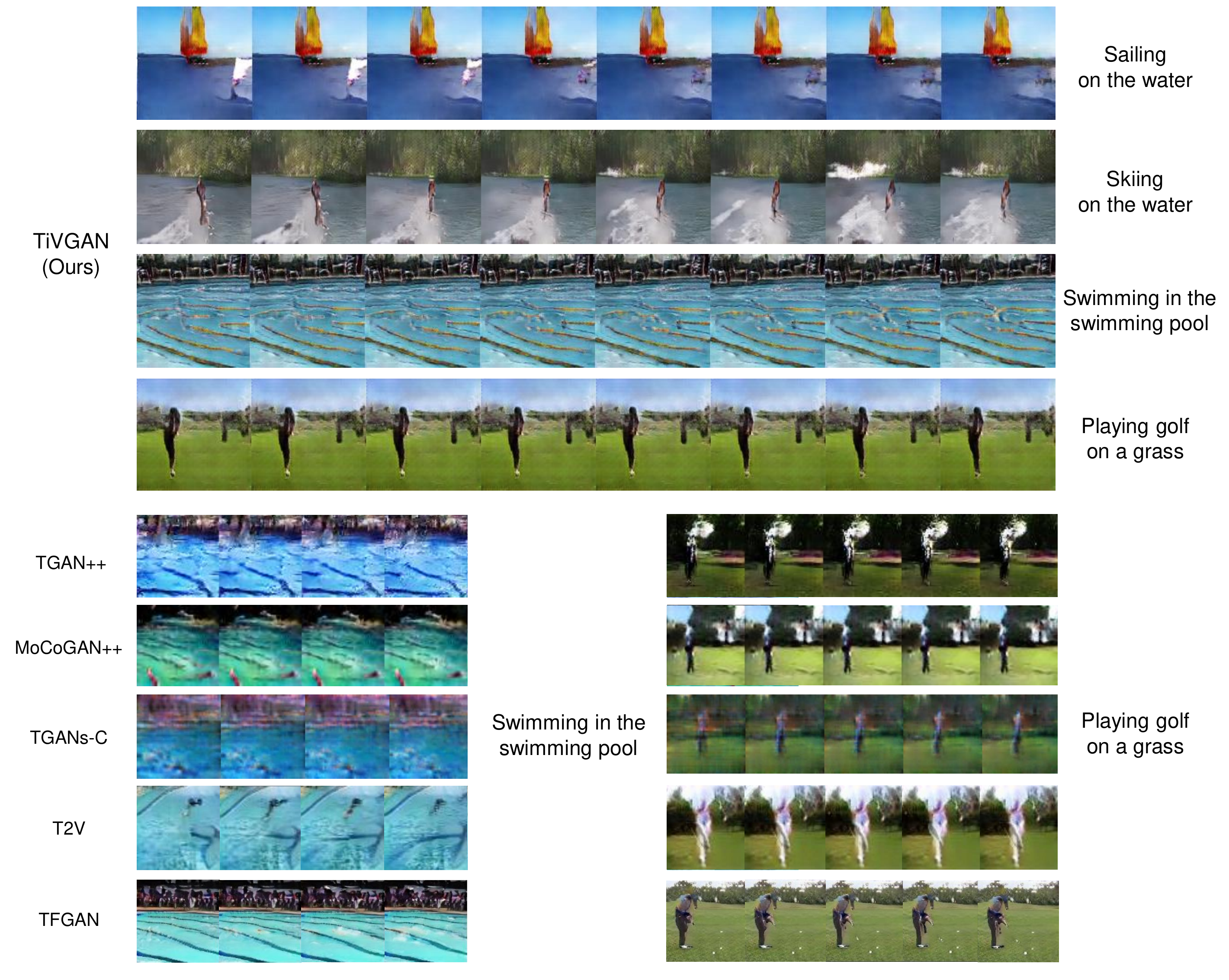}
{\textbf{Example results of text-to-video generation trained on Kinetics dataset. } \label{fig:kinetic_result}}

\noindent \textbf{Quantitative results:} For a quantitative evaluation, we use frechet inception distance(FID)~\cite{heusel2017gans} to measure the quality of the generated images. FID measures the similarity between two image sets. We collect the $200$ video frames generated from each method, and then calculate the FID between each image set with the same number of video frames in the training dataset. Our TiVGAN shows the best performance as shown in Table~\ref{table:KTH}.

\begin{table}[t]
	\begin{center}
		\caption{\textbf{FID score for models trained on the KTH dataset.} 
		A lower FID means that the generated images are more similar to the training data.}
		\begin{tabular}{c|c| c }
		    \toprule
			 & without text& with text \\
			\midrule
			TGAN~\cite{saito2017temporal} & 70.10 & 59.59\\
			MoCoGAN~\cite{tulyakov2018mocogan} & 83.07 & 81.85\\
			TGANs-C~\cite{pan2017create} & - & 69.92\\
			TiVGAN & - & \textbf{47.34}\\
			\bottomrule
		\end{tabular}
    \vspace{-6mm}
	\label{table:KTH}
	\end{center}
\end{table}

\subsection{MUG Facial Expression}
MUG is a human facial expression database~\cite{aifanti2010mug}. There are tens of people in the dataset, and each person shows seven types of facial expressions: `happy', `disgust', `sad', `neutral', `surprise', `fear', and `anger'. We reshaped and used $128\times128$ resolution frames, and the models are trained on a total of $1,030$ videos. Result images on the MUG dataset and their given captions are shown in Figure~\ref{fig:mg_qual}. We observe that our model generates sharp images corresponding to the given text.

\begin{table}[t]
	\begin{center}
		\caption{\textbf{Inception score for models trained on MUG.}}

		\begin{tabular}{c|c| c }
		    \toprule
			 & without text & with text \\
			\midrule
			TGAN~\cite{saito2017temporal} & 3.50 & 4.63\\
			MoCoGAN~\cite{tulyakov2018mocogan} & 3.31 & 4.92\\
			TGANs-C~\cite{pan2017create} & - & 4.65 \\
			TiVGAN & - & \textbf{5.34}\\
			\bottomrule
		\end{tabular}
    \vspace{-6mm}
	\label{table:MUGG}
	\end{center}
\end{table}

\noindent \textbf{Quantitative results:} The inception score is used for the quantitative evaluation of the MUG dataset results. It is a measurement proposed by Salimans \etal~\cite{salimans2016improved} that evaluates the quality of a GAN by observing the diversity and classification confidence of its generated images. In this experiment, a simple 5-layers 3D convolutional neural network is used instead of an inception network owing to the limitations of the number of data and classes; each video is classified into seven classes representing human facial expressions. Table~\ref{table:MUGG} shows a comparison of the results using the MUG dataset. Our results showed the highest inception score among the studied methods.

\begin{table}[t]
\small
	\caption{\textbf{Inception score for models trained on Kinetics. }}
	\begin{center}
		\begin{tabular}{c|c |c }
		    \toprule
			 & without text & with text \\
			\midrule
			TGAN~\cite{saito2017temporal} & 3.71& 4.62\\
			MoCoGAN~\cite{tulyakov2018mocogan} & 3.97 & 4.20\\
			TGANs-C~\cite{pan2017create} & - & 4.87 \\
			TiVGAN & - & \textbf{5.55}\\
			\bottomrule
		\end{tabular}
	\end{center}
\vspace*{-2mm}
\vspace*{-2mm}
	\label{table:kinetics}
\end{table}

\subsection{Kinetics}
Kinetics is a large-scale, human-focused video dataset from YouTube~\cite{kay2017kinetics}. The dataset comprises thousands of video URLs covering $600$ human action classes. We used six classes from this dataset: `snow bike', `swimming', `sailing', `golf', `kite surfing', and `water ski', which are similar to those used in previous works ~\cite{li2018video}. We reshaped and used a $128\times128$ frame size, and every model is trained on a total of $3,032$ videos. For Kinetics, the recent text-to-video generation method proposed by Li~\etal(T2V)~\cite{li2018video} is also compared qualitatively. 

The generated qualitative results are shown in Figure~\ref{fig:kinetic_result}. From the results, we can easily see that our method produces a much higher-quality video. Our generated images are much clearer because of their higher resolution, and they can also capture more distinctive features of given text, apart from the resolution differences. 

\begin{table}[t]
\small
	\caption{\textbf{Classification accuracy for generated videos on Kinetics.} }
	\begin{center}
		\begin{tabular}{c|c |c | c|c}
		    \toprule
			 & In-set & T2V~\cite{li2018video} & TFGAN~\cite{balaji2019conditional} & TiVGAN(Ours) \\
			\midrule
			Acc. (\%) & 78.1 & 42.6 & 76.2 & \textbf{77.8}\\
			\bottomrule
		\end{tabular}
	\end{center}
\vspace*{-2mm}
	\label{table: accuracy}
\end{table}

\noindent \textbf{Quantitative results:}  First, the inception score is again used for the evaluation of the Kinetics dataset. Each video is trained on the 6-class classification by using a 5-layers 3D convolutional neural network. Table~\ref{table:kinetics} shows a comparison of the results on the Kinetics dataset. Our result again shows the highest inception score among the compared methods. Next, the video classification accuracy of the generated results is recorded in Table~\ref{table: accuracy} following the settings of previous text-to-video works~\cite{balaji2019conditional, li2018video}. Our TiVGAN achieves the highest performance which is very close to the in-set accuracy.

\Figure[!t](topskip=0pt, botskip=0pt, midskip=0pt)[width=0.99\linewidth]{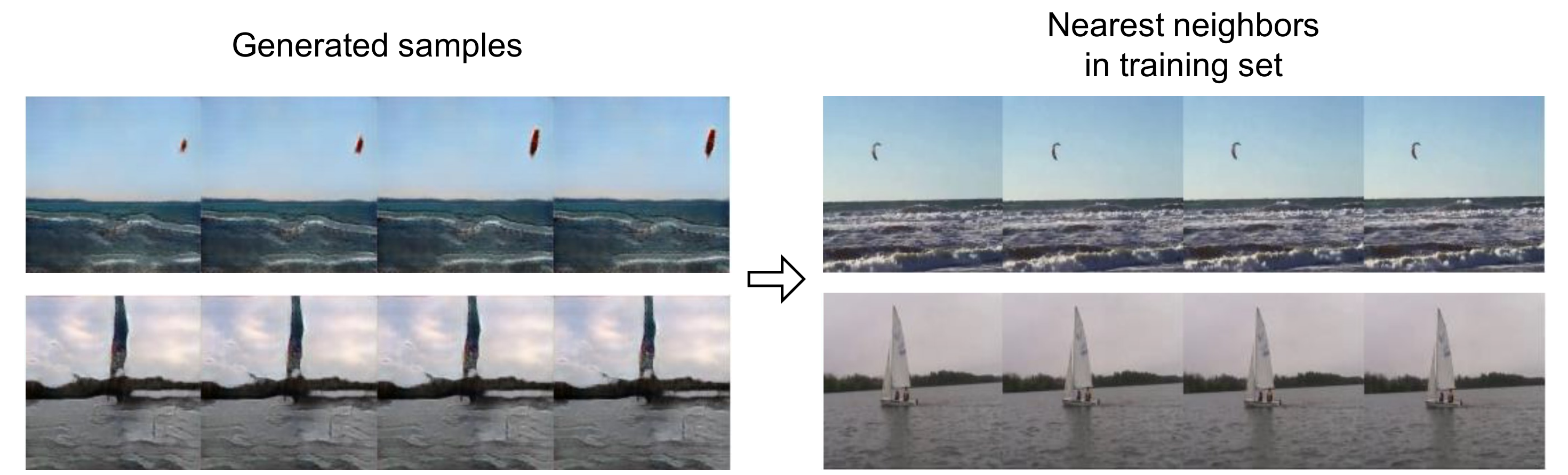}
{\textbf{Ablation study on nearest neighbors. } Left images are generated samples, and right images are corresponding nearest neighbors in training dataset.\label{fig:nearest}}

\section{Ablation studies}

We conduct ablation studies to analyze the effectiveness of the proposed architecture. All experiments are tested on the Kinetics dataset.

\subsection{Nearest Neighbors} To address that our model does not simply memorize the dataset, we present the nearest neighbor image in Figure~\ref{fig:nearest}. We can observe that our generated results are different from the nearest neighbors in the training set.

\subsection{Step-by-Step Generation} 
When we aim to create $2^n$ frames of video, our network starts with generating a single frame ($n=0$) and gradually double the number of images to create ($n=1, 2, 3, 4$). To show the advantage of our step-by-step evolutionary generation framework, we perform an ablation study with various cases. Several steps are omitted in the comparison experiment, but total training time in all cases is same for fair comparison.
\\

\begin{table}[t]
\begin{center}

	\caption{\textbf{Ablation study on step-by-step evolutionary generator for 16 frames video generation.} The first row indicates our final model. The initial text-to-image generation stage is excluded on the second row and the middle two steps are omitted in the third row. In the fourth row, only the last step is performed.}
\begin{tabular}{ccccc|c}
\toprule

\multicolumn{5}{c|}{Number of generated frames}                         & \multirow{2}{*}{Inception Score} \\ \cmidrule{1-5}
\centering
            \small{1}   &  \small{2}   &   \small{4}  &   \small{8}  &  \small{16}   &                   \\ \midrule
            \cmark      &    \cmark    &    \cmark    &    \cmark    &    \cmark    &     \textbf{5.55} \\
                -       &    \cmark    &     \cmark   &   \cmark     &    \cmark    &         5.13      \\
            \cmark      &      -       &     \cmark   &      -       &    \cmark    &         5.47   \\
                 -      &      -       &       -      &      -       &    \cmark    &     5.25 \\ \bottomrule
\end{tabular}
	\label{table:stepgen_abl}
\end{center}
\end{table}

\indent As shown in Table~\ref{table:stepgen_abl}, TiVGAN yields the highest inception score than other experiments. The results of the second row and fourth row demonstrate that initial text-to-image generation plays a significant role in the final step. Also, the result of third row indicates that skipping a few steps decrease the performance of the model. From the experimental results, it can be seen that our step-by-step generation is critical to producing high-quality video.

\Figure[!t](topskip=0pt, botskip=0pt, midskip=0pt)[width=0.99\linewidth]{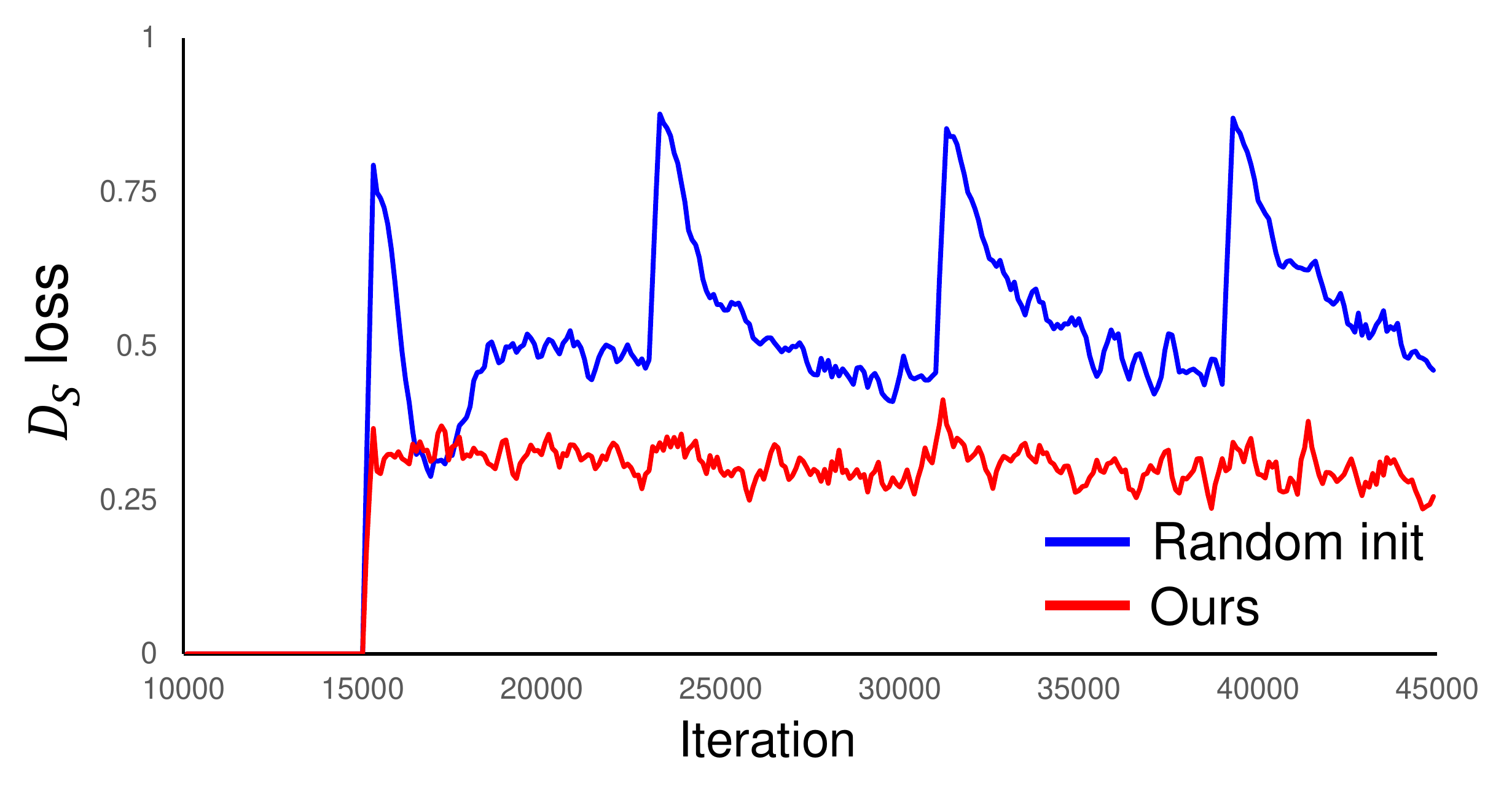}
{\textbf{Ablation study on step discriminator initialization. } Training loss is used as a measure to demonstrate the effectiveness of step-discriminator initialization compared to random initialization. The points where the blue line rises abruptly indicates the time when the step changes.\label{fig:ablation}}

\Figure[!t](topskip=0pt, botskip=0pt, midskip=0pt)[width=0.99\linewidth]{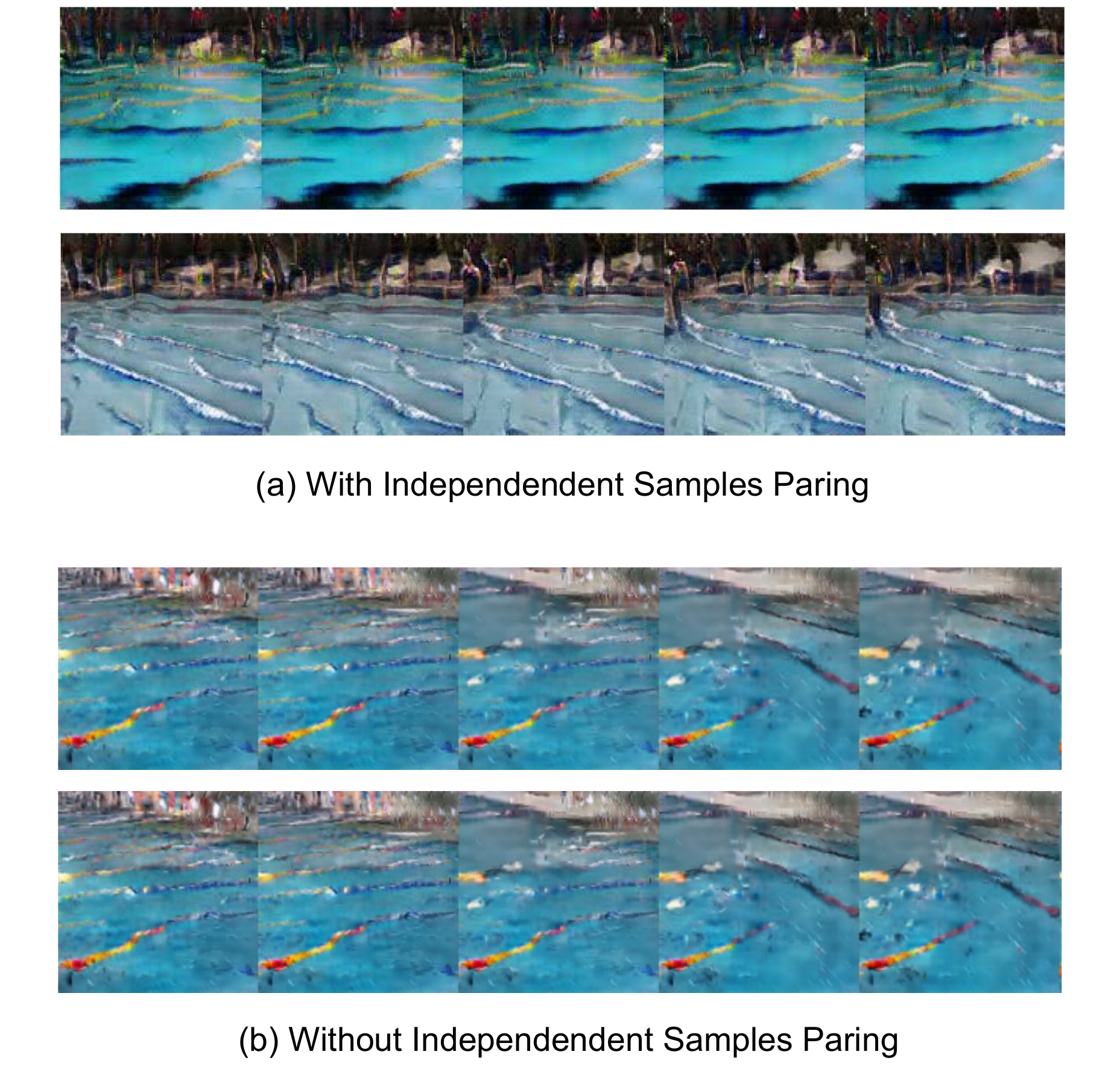}
{\textbf{Ablation study on Independent Samples Pairing.} Two different video clips are independently generated from one text input ‘swimming in the swimming pool’ with different noises. The left (with ISP) generated a completely different but appropriate video following the text description. However, without ISP, two samples are identically generated even with the different input noise. \label{fig:isp}}

\subsection{Step-discriminator Initialization} 
In Sec~\ref{sec:step_discriminator}, we described our step discriminator initialization as our training strategy to enhance the network. In this experiment, we record the training loss of $D_S$ for the cases with and without our initialization. The results are shown in Figure~\ref{fig:ablation}. 
With random noisy initialization, $D_S$ shows an unstable loss graph at the beginning of every step. Since random initialization does not utilize the previous learning, the loss rises rapidly when the discriminator is newly added.
The step-discriminator initialization indicates that $D_S$ is not affected by step change. This means that the model can reliably handle the generation of a larger number of images owing to the suggested initialization for the step-discriminator.

\subsection{Independent Samples Pairing} We employ Independent Samples Pairing to prevent the mode-collapse of the generator. The effects of ISP can be visualized in Figure~\ref{fig:isp}. Without ISP, the generator often produces identical outputs when the same input text with different noise are given. However, we verify that our network generates various videos when the same text description and different random noise are given.

\section{Conclusion}

In this paper, we proposed a new effective learning paradigm for text-to-video generation. Beginning with the creation of a single image, our network evolves progressively to synthesize a video clip of a desired length. Additionally, several techniques were used for stabilizing the training. Experimental results on the KTH, MUG, and Kinetics datasets support that our model can accomplish the given task under various situations. Conditional video generation is still a less explored field, but we believe it will be actively researched in the near future. We hope that our work will invite more interest in this field.

\bibliographystyle{IEEEtran}
\bibliography{egbib}

\begin{IEEEbiography}[{\includegraphics[width=1in,height=1.25in,clip,keepaspectratio]{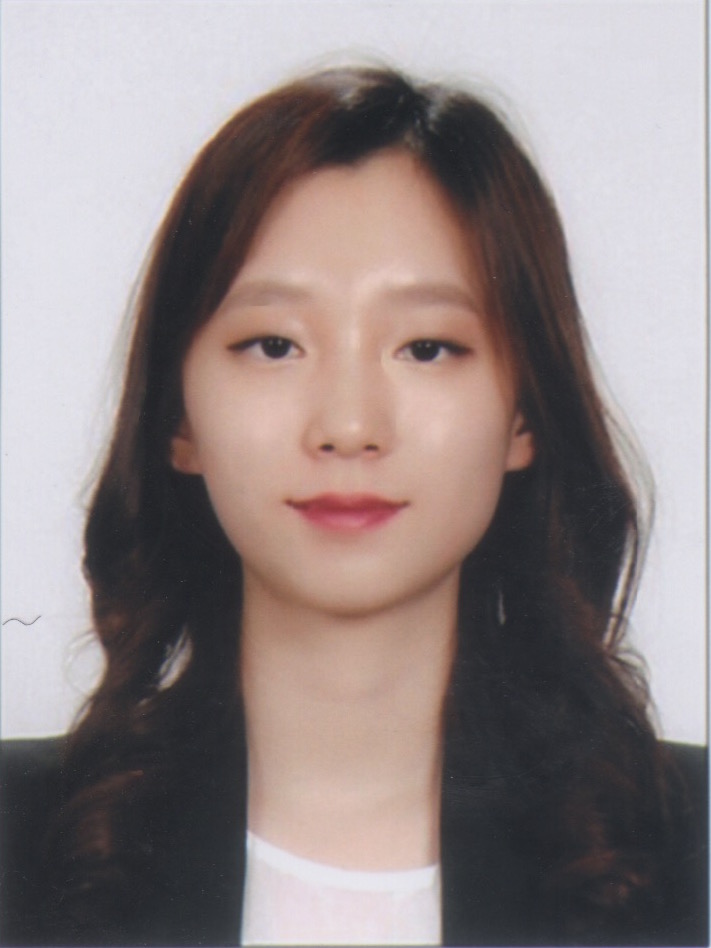}}]{Doyeon Kim} received the B.S. degree in computer science from Korea University, Seoul, South Korea, in 2016, and the M.S. degree in robotics program from Korea Advanced Institute of Science and Technology (KAIST), Daejeon, South Korea, in 2018. She is currently pursuing the Ph.D. degree of electrical engineering with KAIST. Her research interests include computer vision, deep learning, and machine learning.

\end{IEEEbiography}

\begin{IEEEbiography}[{\includegraphics[width=1in,height=1.25in,clip,keepaspectratio]{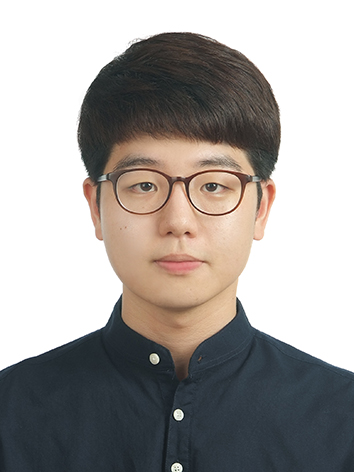}}]{donggyu joo} received the B.S. and M.S. degrees in electrical engineering from Korea Advanced Institute of Science and Technology (KAIST), Daejeon, South Korea, in 2016 and 2018, respectively. He is currently pursuing the Ph.D. degree of electrical engineering with KAIST. His research interests include computer vision, deep learning, and machine learning. 
\end{IEEEbiography}

\begin{IEEEbiography}[{\includegraphics[width=1in,height=1.25in,clip,keepaspectratio]{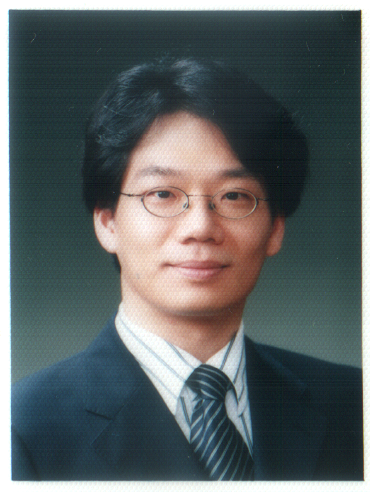}}]{Junmo Kim} (S’01-M’05) received the B.S. degree from Seoul National University, Seoul, Korea, in 1998, and the M.S. and Ph.D. degrees from the Massachusetts Institute of Technology (MIT), Cambridge, in 2000 and 2005, respectively. From 2005 to 2009, he was with the Samsung Advanced Institute of Technology (SAIT), Korea, as a Research Staff Member. He joined the faculty of KAIST in 2009, where he is currently an Associate Professor of electrical engineering. His research interests are in image processing, computer vision, statistical signal processing, machine learning, and information theory.
\end{IEEEbiography}

\EOD

\end{document}